\documentclass{article}
\usepackage{ijcai18}

\usepackage{times}
\usepackage{xcolor}
\usepackage{soul}
\usepackage[utf8]{inputenc}
\usepackage[small]{caption}
\usepackage{graphicx}
\usepackage{color}

\usepackage{url}

\title{Classifying action correctness in physical rehabilitation exercises}

\author{
Alina Miron, 
Crina Grosan
\\ 
Brunel University London \\
alina.miron@brunel.ac.uk,
crina.grosan@brunel.ac.uk
}

\begin{document}

\maketitle

\begin{abstract}
 The work in this paper focuses on the role of machine learning in assessing the correctness of a human motion or action. This task proves to be more challenging than the gesture and action recognition ones. We will demonstrate, through a set of experiments on a recent dataset, that machine learning algorithms can produce good results for certain actions, but can also fall into the trap of classifying an incorrect execution of an action as a correct execution of another action. 
 
\end{abstract}

\section{Introduction}

Analyzing human motion has been an intensely studied problem in the computer vision community. While most works focus on the challenging task of action detection and recognition, there is still limited work in the domain of human motion quality assessment from a functional point of view. 
Several potential applications for this exists, including heath care for patient rehabilitation  and sports for athlete performance improvement \cite{pirsiavash2014assessing}.

Assessing the quality of human motion/actions is a difficult problem. Human experts such as coaches, physiotherapists, or doctors have been trained extensively to discover the rules required to assess different types of motions.

In this work, we concentrate on motion quality assessment from a correctness perspective, using machine learning methods. We want to see whether machine learning methods could easily classify an action, from a set of various actions correctly and incorrectly executed, as valid or not, in a binary manner. 

\section{Related work}

The task of everyday activity recognition \cite{pirsiavash2012detecting}\cite{shahroudy2016ntu} and action recognition \cite{idrees2017thumos} have been discussed in several papers, with a large number of datasets being publicly available \cite{escalera2017challenges}. 

There are a few articles where the task of \textit{action correctness} is approached from different perspectives. Some approaches use either accelerometer sensors, others rely on depth or colour cameras.

\cite{ebert2017qualitative} tackles the task of a qualitative assessment of human motion through the use of an accelerometer and assigning a quality class label to each motion. Other authors, such as \cite{parisi2016human}, focus on computing how much a performed movement recorded using a depth camera matches the correct continuation of a learned sequence. \cite{paiement2014online} uses the recorded gait movement of six healthy subjects going up the stairs to train a model. The model's ability to detect the abnormalities is tested on 6 other patients with 3 types of simulated knee injuries. 

Action correctness is also very similar to  action completeness \cite{heidarivincheh2016beyond}. In this context, an action is considered completed if the action goal was achieved: i.e. \textit{the drinking action is completed when one actually consumes a beverage from a cup}. The authors have used six types of actions to test the completeness, which involves the interaction with different objects. In this article, we do not aim for action completeness, but instead we focus on the specific task of how correct an action is actually performed.

For the action correctness task, the only  publicly available dataset  that we found is UI-PRMD, proposed by  \cite{vakanski2018data}. They have recorded 10 subjects performing 10 types of actions, with each action being performed in an optimal and non-optimal way. The dataset was not recorded with a particular type of injury in mind, but focuses instead on healthy subjects performing a few types of exercises. 

At the moment, no public baseline benchmark has been published on the UI-PRMD dataset. Nevertheless, we are studying the feasibility of training an action correctness model on this dataset. We construct a binary classifier for every type of exercise, with the purpose of differentiating between a correctly and wrongly executed action. Different subjects might perform the non-optimal movement in several ways. For example, for the \textit{"Deep squat"} exercise, the non-optimal movement is defined by \cite{vakanski2018data} as \textit{"Subject does not maintain upright trunk posture, unable to squat past parallel, demonstrates knee values collapse or trunk flexion greater than 30$^{\circ}$"}. This definition allows a certain degree of subjectivity in assessing the correctness of the movement. 

\begin{figure*}[h!]
  \includegraphics[width=\textwidth]{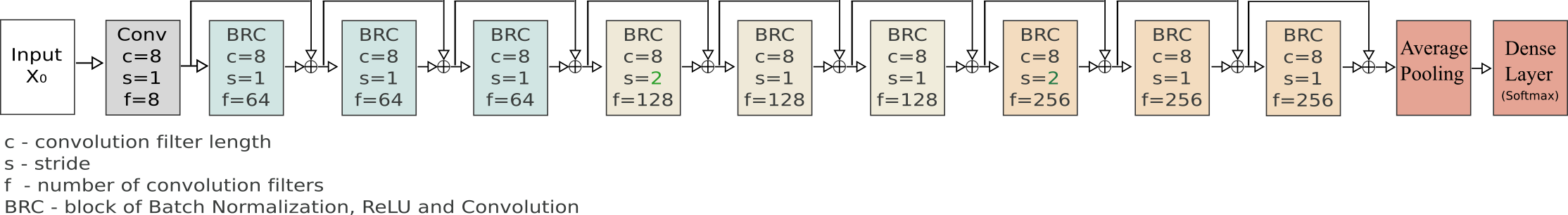}
  \caption{The structure of the Res-TCN, where \textbf{BRC} stands for \textbf{B}atch Normalization,  \textbf{R}eLU and \textbf{C}onvolution}
  \label{fig:tcnstructure}
\end{figure*}

Instead of using hand crafted features for every type of action, our purpose is to use a machine learning system to learn what makes a movement non-optimal. We use the Temporal Convolutional Neural Network (Res-TCN) proposed by \cite{kim2017interpretable}. This classifier is one of the top performing methods on a large scale action recognition dataset \cite{shahroudy2016ntu}, giving slightly better performance than the STA-LSTM used by \cite{song2017end}.

\section{Machine learning for action correctness}

The results presented in this section are obtained with Convolutional Neural Networks. It is worth mentioning that a few other standard machine learning methods have been tested, including Support Vector Machines and Random Forests, but the classification accuracy was not significantly higher that 50\%, which is close to a random decision.

\subsection{Overview of the Res-TCN classifier}

In this section we provide a short overview of the structure of the network proposed by \cite{kim2017interpretable}. 

Figure \ref{fig:tcnstructure} depicts the structure of the Res-TCN network. The input $X_0$ to the network is the concatenated skeleton features from every  frame of the video sequence. This is followed by a first convolution with the convolution filter length $c$ of eight, a stride $s$ of one and number of filters $f$ of eight.

The following nine blocks are \textit{Residual Units} introduced by \cite{he2016deep} and consist of batch normalization, ReLU and convolution operations, with the number of filters of the convolution increasing from 64 to 128 and 256. After the last layer a global average pooling is used across the entire temporal sequence. A final softmax layer with the number of neurons equal to the number of classes is used for classification.

The advantage of the Res-TCN architecture over recurrent structures like LSTM alternatives is possible model interpretability as shown by \cite{kim2017interpretable}.

\subsection{Model parameters}

\begin{figure*} [h!]
\includegraphics[width=\linewidth]{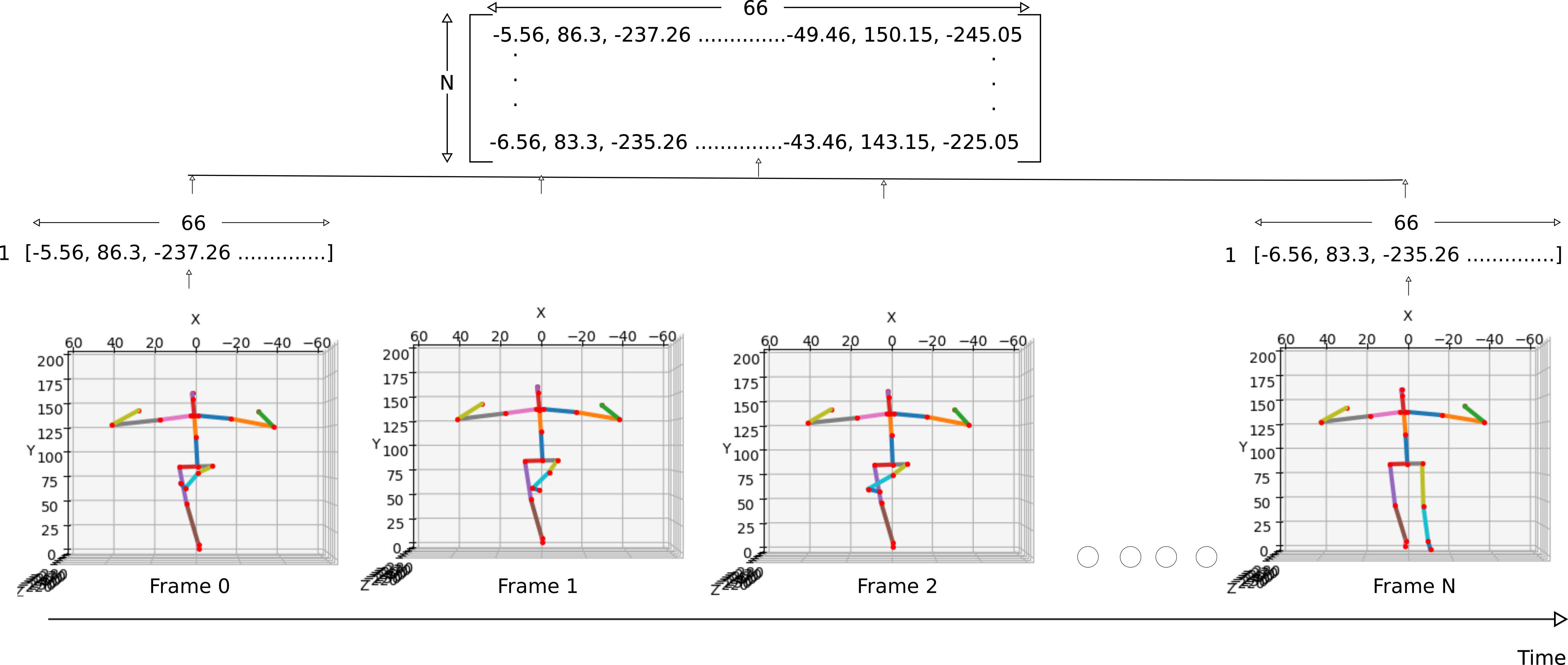}
\caption{The raw feature extraction for a single sample from the UI-PRMD dataset}
\label{fig:X0}
\end{figure*}

In \cite{kim2017interpretable}, the authors are using as input to the TCN the raw 3D skeleton points. We have used a similar setup, but have also tested the system performance when it receives as input the angles between different joints. For 3D skeleton points setup, we take the computed (X, Y, Z) values of each skeleton joint and concatenate all values to form a skeleton feature. A skeleton feature per frame is a 66 dimensional vector obtained by multiplying the number of joints (which is 22) with the data per point (which is 3)  as it can be seen in Figure \ref{fig:X0}.

One disadvantage of a TCN architecture over other types of architectures like LSTM is the fact that the data size has to be consistent over all input examples. The way we overcome this limitation is by finding the maximum video length across all segmented movements and use zero padding for the  2D feature-array. 

For the Res-TCN parameters, we used the same configuration as proposed by \cite{kim2017interpretable}: stochastic gradient descent with nesterov acceleration with a momentum of 0.9, all convolution layers have applied a $L1$ regularizer with a weight of $10^{-4}$, and to prevent overfitting a dropout of 0.5 is applied after every ReLU.

The model is trained for 500 epochs, and we use a batch size of 128.

\subsection{Data and experiments}

As mentioned above, we use the UI-PRMD dataset for our machine learning investigation. The data has been recorded using both a Kinect camera and a Vicon optical tracker \cite{vicon}.  The Vicon optical tracker is a system designed for capturing human motion with high accuracy and consists of eight high speed cameras that track a set of retroreflective markers. We have focused just on the data recorded with the Kinect, due to the low cost of the sensor compared with Vicon.

UI-PRMD data consists of 10 movements: deep squat, hurdle step, inline lunge, side lunge, sit to stand, standing active straight leg raise, standing shoulder abduction, standing shoulder extension, standing shoulder internal-external rotation, and standing shoulder scaption. Each movement is repeated 10 times by each of the 10 individuals recorded. 

The Kinect data is presented in the form of 22 YXZ triplets of Euler angle and positions. The values for the waist joint are given in absolute coordinates, while the values of the rest of the joints are given in relative coordinates with respect to the parent joint \cite{vakanski2018data}. Based on the local angle and position data we have computed the transformation matrix in order to obtain the absolute joint coordinates as 3D points. 

We followed a cross-subject evaluation splitting the data into training, validation and testing. For the training phase we have used 6 persons, 3 were used in validation and 1 for testing. We have used a 10-fold cross validation in order to validate our model, every time using a different person in the testing set.

The authors of \cite{shahroudy2016ntu} have defined the testing protocol for  NTU RGB+D dataset as a  training/testing split. We find that due to the smaller size of UI-PRMD data, a protocol training/validation/testing to be more appropriate. We use the validation set to avoid model overfitting on the training set. We save the model which generalized best, i.e. obtained the best accuracy, on the validation set, and that model is used on the testing set.

\subsection{Results and discussions}

\begin{figure}[h!]
\includegraphics[width=\linewidth]{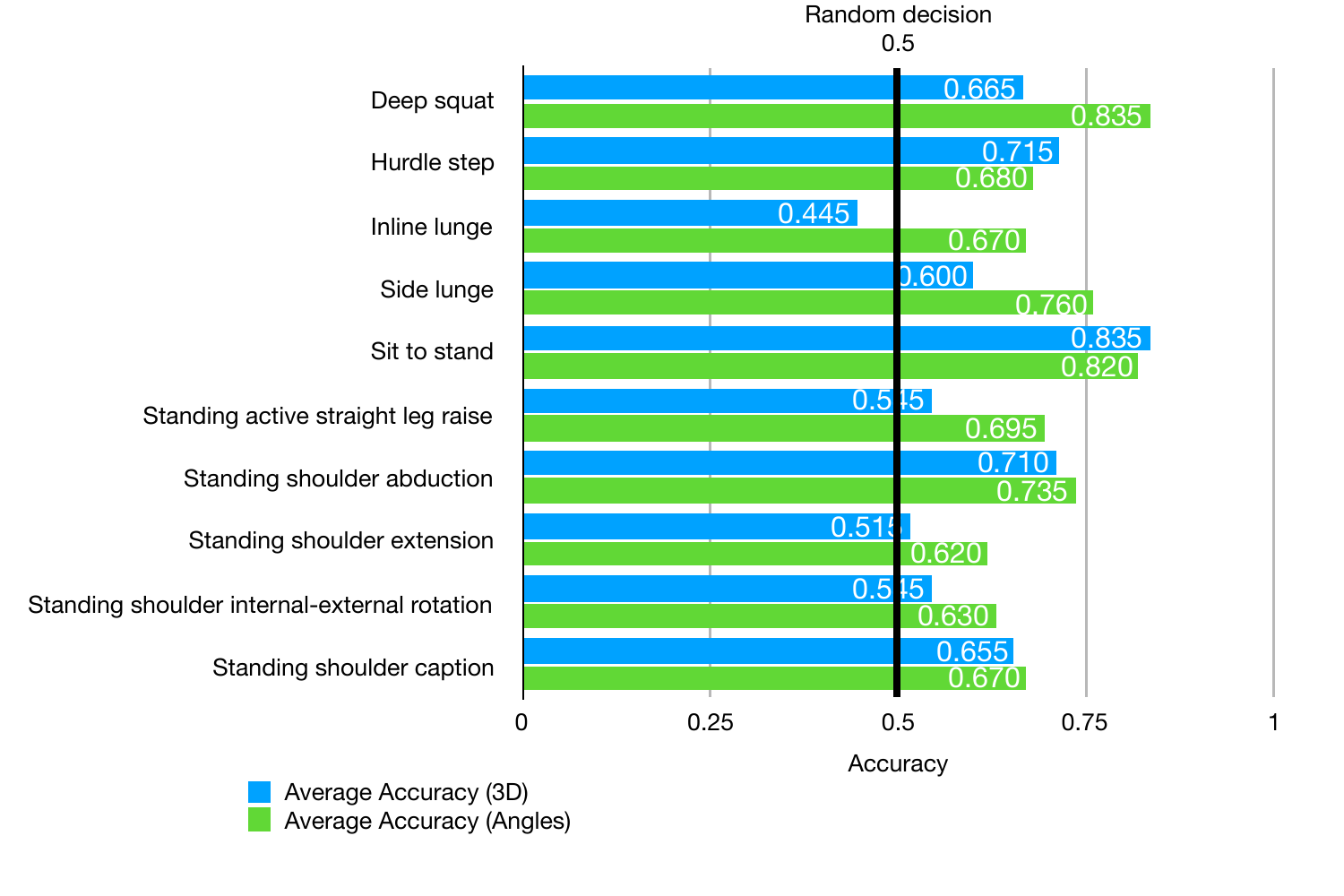}
\caption{Average accuracy over 10-folds for every action}
\label{fig:results}
\end{figure}

For every movement type, we have trained a different model with the explicit purpose of differentiating between optimal and non-optimal movements.  Figure \ref{fig:results} presents the average results for every movement type, by using as input for the neural network the absolute 3D point position or the relative joint angles. 

When using the 3D points, on average the accuracy is 62.3\%, while when using the relative joint angles is 71.2\%.

The types of movements for which the non-optimal movement was consistently detected across subjects are: \textit{deep squat} with 83.5\% accuracy for relative joint angles and \textit{sit to stand} with 82\% accuracy. The movements that proved to be the most difficult are \textit{standing shoulder extension} with 62\% accuracy and \textit{standing shoulder internal-external rotation} with 63\% accuracy.

\begin{figure}[h!]
\includegraphics[width=\linewidth]{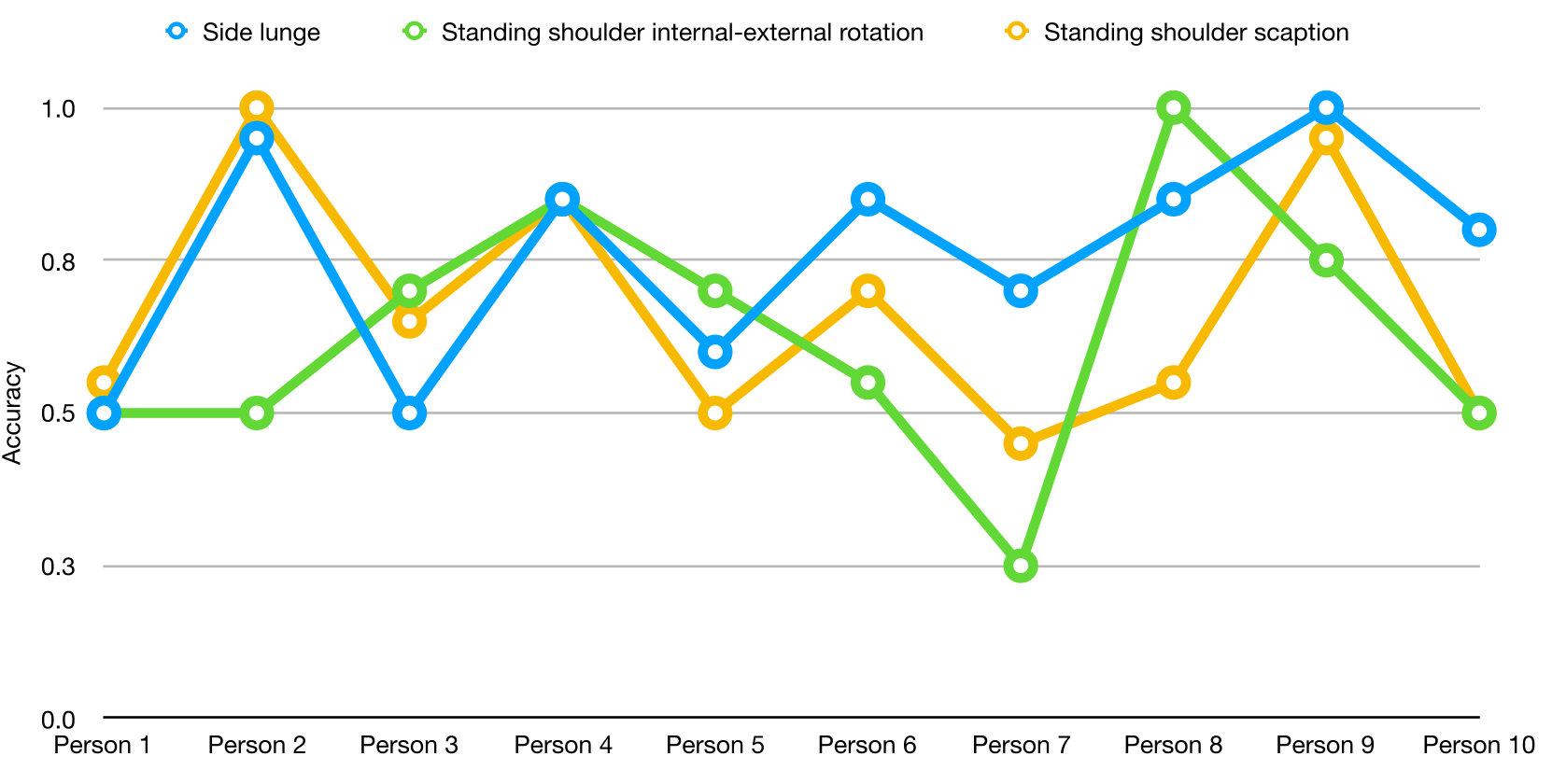}
\caption{Accuracy for side lunge, standing shoulder internal-external rotation and standing shoulder scaption for each test case}
\label{fig:results3Gestures}
\end{figure}

In figure \ref{fig:results3Gestures} we present a detailed view of the classifier performance for 3 actions (side lunge, standing shoulder internal-external rotation and standing shoulder scaption) and all subjects in the dataset. For example, we obtained the accuracy of ~50\% for \textit{Person 1}  for \textit{side lunge} by training a binary classifier where the training set consisted of \textit{Person 2} to \textit{7}, the validation set of \textit{Person 8} and \textit{9}, and the test set of \textit{Person 1}.  An accuracy of 50\% is equivalent to a random decision level, therefore the classifier wasn't able to generalize for this particular subject. On the other hand, for \textit{Person 2} and \textit{8} the classifier achieved a perfect accuracy for certain types of actions.

One reason for this discrepancy of performance between different subjects might be given by the way the actions were actually performed. For example, for the three analyzed actions, side lunge, standing shoulder internal-external rotation and standing shoulder scaption, the motion is performed using the right side of the body (right leg, right shoulder) by most of the subjects. The exception are Person 7 and Person 10 which perform the motion using their left side of the body.

We have also  trained a general model that classifies for any input movement if it is performed correctly or not and it obtained an average accuracy of 63.3\% when using joint angles as input data. This is much lower than the 71.2\% accuracy obtained when training specialized models for every type of movement. 

\section{Conclusion}
The work presented here is an initial investigation of the applicability of machine learning to human motion quality assessment. In particular, we looked into the task of training a model to recognize an action or a movement as correct or not. Our results show a high variability of results for different action types. Angle features seem to be more relevant than using raw 3D joint positions. More work has to be done in identifying, adapting, and improving certain machine learning methods, such as the convolutional neural networks which prove efficient at least for some classes of actions.

The results shown that the actions which involve a movement symmetry of the body, like sit to stand and deep squat, were easier to train. For certain classes of actions, the variability of movement is too high. Therefore, we believe that a data augmentation process might help especially with these action types. As future work, we plan to augment the training set with simulated data, by translating the motion performed by one side of the body to the other one.


\end{document}